\documentclass[letterpaper]{article} 
\usepackage{aaai24}  
\usepackage{times}  
\usepackage{helvet}  
\usepackage{courier}  
\usepackage[hyphens]{url}  
\usepackage{graphicx} 
\usepackage{natbib}  
\usepackage{caption} 
\usepackage{algorithm}
\usepackage{algorithmic}
\usepackage{multirow}
\usepackage{booktabs}
\usepackage{pifont}
\usepackage{amsmath}
\usepackage{amsfonts}
\usepackage[pagebackref,breaklinks,colorlinks]{hyperref}

\usepackage{newfloat}
\usepackage{listings}
\DeclareCaptionStyle{ruled}{labelfont=normalfont,labelsep=colon,strut=off} 
\floatstyle{ruled}
\newfloat{listing}{tb}{lst}{}
\floatname{listing}{Listing}

\title{VQCNIR: Clearer Night Image Restoration with Vector-Quantized Codebook}
\author{ 
    Wenbin Zou,\textsuperscript{\rm 1}
    Hongxia Gao$\dagger$,\textsuperscript{\rm 1,2}\thanks{Corresponding author.}
    Tian Ye,\textsuperscript{\rm 3}
    Liang Chen,\textsuperscript{\rm 4} \\
    Weipeng Yang,\textsuperscript{\rm 1}
    Shasha Huang,\textsuperscript{\rm 1}
    Hongshen Chen,\textsuperscript{\rm 1}
    Sixiang Chen\textsuperscript{\rm 3}
}
\affiliations{
    \textsuperscript{\rm 1}The School of Automation Science and Engineering, South China University of Technology,\\
    \textsuperscript{\rm 2}Research Center for Brain-Computer Interface, Pazhou Laboratory, Guangzhou,\\
    \textsuperscript{\rm 3}The Hong Kong University of Science and Technology, Guangzhou,\\
    \textsuperscript{\rm 4} College of Photonic and Electronic Engineering, Fujian Normal University

    $\dagger$Email: hxgao@scut.edu.cn
    \vspace{25pt}
}

\usepackage{bibentry}

\begin{document}

\maketitle

\begin{abstract}
Night photography often struggles with challenges like low light and blurring, stemming from dark environments and prolonged exposures. Current methods either disregard priors and directly fitting end-to-end networks, leading to inconsistent illumination, or rely on unreliable handcrafted priors to constrain the network, thereby bringing the greater error to the final result. We believe in the strength of data-driven high-quality priors and strive to offer a reliable and consistent prior, circumventing the restrictions of manual priors.

In this paper, we propose Clearer Night Image Restoration with Vector-Quantized Codebook (VQCNIR) to achieve remarkable and consistent restoration outcomes on real-world and synthetic benchmarks. To ensure the faithful restoration of details and illumination, we propose the incorporation of two essential modules: the Adaptive Illumination Enhancement Module (AIEM) and the Deformable Bi-directional Cross-Attention (DBCA) module. The AIEM leverages the inter-channel correlation of features to dynamically maintain illumination consistency between degraded features and high-quality codebook features. Meanwhile, the DBCA module effectively integrates texture and structural information through bi-directional cross-attention and deformable convolution, resulting in enhanced fine-grained detail and structural fidelity across parallel decoders.

Extensive experiments validate the remarkable benefits of VQCNIR in enhancing image quality under low-light conditions, showcasing its state-of-the-art performance on both synthetic and real-world datasets. The code is available at https://github.com/AlexZou14/VQCNIR.
\end{abstract}

\section{Introduction}

To obtain reliable images in night scenes, long exposure is often used to allow more available light to illuminate the image. However, images captured in this way still suffer from low visibility and color distortion issues. Moreover, long exposure is susceptible to external scene disturbances, such as camera shake and dynamic scenes, which can cause motion blur and noise in the images \shortcite{LEDNet}. Therefore, night images often exhibit complex degradation problems~\shortcite{LLFlow,liu2022nighttime,liu2023nighthazeformer} such as low illumination and blur, making the recovery of high-quality images with realistic texture and normal lighting conditions extremely challenging.

\begin{figure}
	\includegraphics[width=8.4cm]{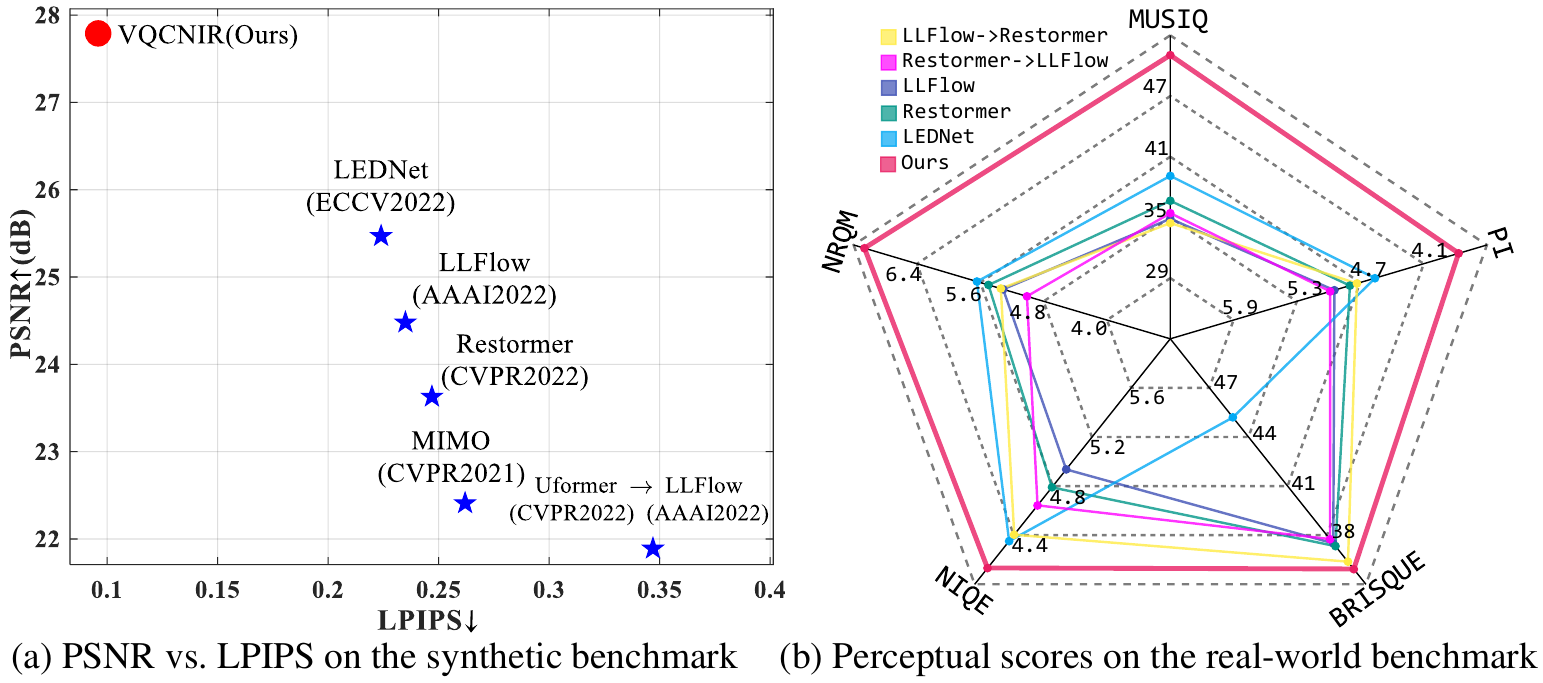}
	\caption{Quantitative comparisons with state-of-the-art methods. (a) PSNR and LPIPS results on the LOL-Blur dataset. (b) Results for five perceptual metrics on the Real-LOL-Blur dataset. For PSNR, MUSIQ \shortcite{MUSIQ}, and NRQM \shortcite{NRQM} higher is better, while lower is better for LPIPS \shortcite{LPIPS}, NIQE \shortcite{NIQE}, BRISQUE \shortcite{BRISQUE}, and PI \shortcite{PI}.}
	\label{cpreal1}
\end{figure}

With the great success of deep learning methods~\shortcite{ye2022perceiving,chen2023uncertainty,chen2023cplformer,chen2023msp,ye2023sequential,chen2023sparse} in image restoration and video restoration, numerous deep learning-based algorithms have been proposed to tackle this challenging task. Currently, most researchers only consider the low illumination problem in night images and have proposed numerous low-light image enhancement (LLIE) methods \shortcite{LLNet, RetinexNet, Kind, Zero-DCE, Zero-DCE++, DeepUPE, UTVNet}. Although these LLIE methods can produce visually pleasing results, their generalization ability is limited in real night scenes. This is mainly attributed to the fact that LLIE methods focus primarily on enhancing image luminance and reducing noise while ignoring the spatial degradation caused by blur that leads to ineffective recovery of sharp images. An intuitive idea is to combine image deblurring methods with LLIE methods to address this problem. However, most existing deblurring methods \shortcite{LOLBlur_ref_5, LOLBlur_ref_7, LOLBlur_ref_48, restormer, uformer} are trained on datasets captured under normal illumination conditions, which makes them not suitable for night image deblurring. In particular, due to the poor visibility in dark regions of night images, these methods may fail to effectively capture motion blur cues, resulting in unsatisfactory deblurring performance. Therefore, simply cascading LLIE and deblurring methods do not produce satisfactory recovery results. To better handle the joint degradation process of low illumination and blur, Zhou et al. \shortcite{LEDNet} first proposed a LOL-Blur dataset and an end-to-end encoder-decoder network called LEDNet. LEDNet can achieve high performance on the synthetic LOL-Blur dataset. However, its generalization ability in real scenes is still limited.

The aforementioned night restoration methods have difficulties in recovering correct textures and reliable illuminations from low-quality night images. \textit{\textbf{This is due to the lack of stable and reliable priors, as most existing priors are generated from low-quality images.}} For instance, Retinex-based techniques \shortcite{RetinexNet,Kind,KinD++} employ illumination estimation through the decomposition of low-quality images, while blur kernels are estimated using the same degraded inputs. However, the biased estimation of priors leads to cumulative errors in the final outcomes.

Therefore, we introduce the vector quantization (VQ) codebook as a credible and reliable external feature library to provide high-quality priors for purely data-driven image restoration, instead of relying on vulnerable handcrafted priors.

The VQ codebook is an implicit prior generated by a VQGAN \shortcite{VQGAN} and trained on a vast corpus of high-fidelity clean images. Hence, a well-trained VQ codebook can provide comprehensive, high-quality priors for complex degraded images, effectively addressing complex degradation. \textit{\textbf{Furthermore, inconsistent illumination and incorrect matching between the degradation features of night images and the pristine features in the VQ codebook can lead to unsatisfactory visual effects when directly reconstructing using the codebook.}} It may even amplify blur and produce artifacts in restored images. Hence, the pivotal step towards harnessing codebook priors for the restoration of night-blurred images lies in precisely aligning the high-quality codebook features.

In this paper, we propose a novel method called \textbf{Clearer Night Image Restoration with Vector-Quantized Codebook (VQCNIR)} for night image restoration. To address the aforementioned key considerations, our proposed VQCNIR incorporates two purpose-built modules. Specifically, we design the \textbf{Adaptive Illumination Enhancement Module (AIEM)} that leverages inter-channel correlations of features to estimate curve parameters and adaptively enhances illumination in the features. This effectively addresses inconsistent illumination between degraded features and high-quality VQ codebook features. To ameliorate feature mismatch between degraded and high-quality features, we propose a parallel decoder integrating \textbf{Deformable Bi-directional Cross-Attention (DBCA)}. This parallel design effectively incorporates high-quality codebook features while efficiently fusing texture and structural information from the parallel encoder. Our proposed DBCA performs context modeling between high and low-quality features, adaptively fusing them to gradually recover fine details that enhance overall quality. As depicted in Figure \ref{cpreal1}, our method not only achieves superior performance on synthetic data, but also generalizes well to real-world scenes. Extensive experiments on publicly available datasets demonstrate that our method surpasses existing state-of-the-art methods on both distortion and perceptual metrics.

Our key contributions are summarized as follows:
\begin{itemize}
    \item We propose VQCNIR, a new framework that formulates night image restoration as a matching and fusion problem between degraded and high-quality features by introducing a high-quality codebook prior. This addresses limitations of previous methods that rely solely on low-quality inputs, and achieves superior performance.
    \item We propose an adaptive illumination enhancement module that utilizes the inter-channel dependency to estimate curve parameters. This effectively addresses the inconsistency of illumination between the degraded features and high-quality VQ codebook features.
    \item We further propose a deformable bi-directional cross-attention, which utilizes a bi-directional cross-attention mechanism and deformable convolution to address the misalignment issue between features from the parallel decoder and restore the more accurate texture details.
\end{itemize}

\begin{figure*}
\setlength{\abovecaptionskip}{0.1cm} 
\setlength{\belowcaptionskip}{-0.5cm}
	\centering
	\includegraphics[width=16cm]{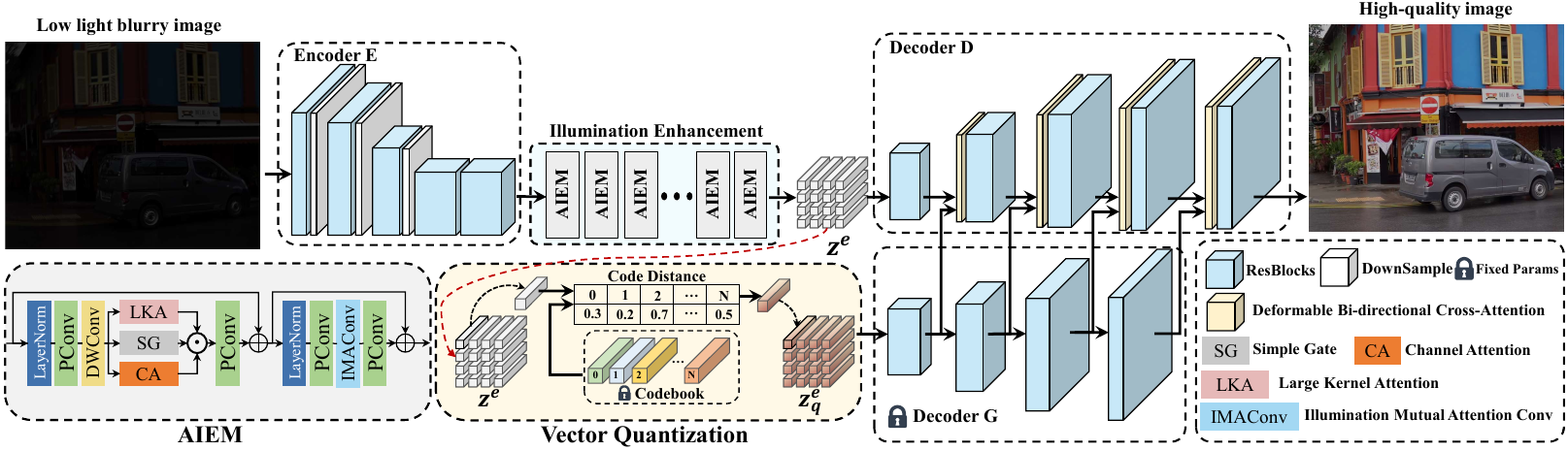}\\
	\caption{The framework of the proposed VQCNIR. It consists of an encoder, some adaptive illumination enhancement modules (AIEM), and a parallel decoder with a deformable bi-directional cross-attention (DBCA), allowing the network effectively to exploit high-quality codebook prior information.}
	\label{VQCNIR}
\end{figure*}

\section{Related Work}
\subsection{Image Deblurring}
Recent advances in deep learning techniques have greatly impacted the field of computer vision. A large number of deep learning methods have been proposed for both single image and video deblurring tasks \shortcite{LOLBlur_ref_11, LOLBlur_ref_30, LOLBlur_ref_36, LOLBlur_ref_48, DeblurGANv2, LOLBlur_ref_51, LOLBlur_ref_5}, and have demonstrated superior performance. With the introduction of large training datasets for deblurring tasks \shortcite{dataset1, LOLBlur_ref_30, dataset3}, many researchers \shortcite{dataset1, dataset3} have adopted end-to-end networks to directly recover clear images. Despite the fact that end-to-end methods outperform traditional approaches, they may not be effective in cases with severe blurring. To improve network performance, some methods \shortcite{LOLBlur_ref_30, LOLBlur_ref_36, LOLBlur_ref_7} use multi-scale architectures to enhance deblurring at different scales. 

However, the limited ability of these methods to capture the correct blur cues in low-light conditions, particularly in dark areas, has hindered their effectiveness in handling low-light blurred images. To tackle this issue, Zhou et al. \shortcite{LEDNet} introduce a night image blurring dataset and develop an end-to-end UNet architecture that incorporates a learnable non-linear layer to effectively enhance dark regions without overexposing other areas. Although the method achieves good performance, it is not capable of generalizing enough in real-world scenes, indicating that there is still room for improvement in handling low-light blurred images.

\subsection{Low-light Image Enhancement}
Recent years have witnessed the impressive success of deep learning-based low-light image enhancement (LLIE) since the first pioneering work \shortcite{LLIE}. Many end-to-end methods \shortcite{LLNet, DeepUPE} have been proposed for enhancing image illumination using an encoder-decoder framework. To further improve the performance of LLIEs, researchers have developed deep Retinex-based methods \shortcite{RetinexNet, Kind, UTVNet} inspired by Retinex theory, which employs dedicated sub-networks to enhance the illuminance and reflectance components and achieve better recovery performance. However, such methods have limitations, as the enhancement results strongly depend on the characteristics of the training data. To improve the generalization ability of the network, researchers \shortcite{Zero-DCE, Zero-DCE++, EnlightenGAN} propose a number of unsupervised methods. For example, Jiang et al \shortcite{EnlightenGAN} introduced self-regularization and unpaired training into LLIE with EnlightenGAN. Additionally, Guo et al \shortcite{Zero-DCE} propose a fast and flexible method for estimating image enhancement depth curves that do not require any normal illumination reference images during the training process.

\subsection{Verctor-Quantized Codebook}
VQVAE \shortcite{VQVAE} is the first to introduce vector quantization (VQ) techniques into an autoencoder-based generative model to achieve superior image generation results. Specifically, the encoded latent variables are quantized to their nearest neighbors in a learnable codebook, and the resulting quantized latent variables are used to reconstruct the data samples. Building upon VQVAE, subsequent work has proposed various improvements to codebook learning. For instance, VQGAN \shortcite{VQGAN} utilizes generative adversarial learning and refined codebook learning to further enhance the perceptual quality of reconstructed images. The well-trained codebook can serve as a high-quality prior that can be leveraged for various image restoration tasks such as image super-resolution and face restoration. To this end, Chen et al. \shortcite{FeMaSR} introduce a VQ codebook prior for blind image super-resolution, which matches distorted LR image features with distortion-free HR features from a pre-trained HR prior. Furthermore, Gu et al. \shortcite{VQFR} explore the impact of internal codebook properties on reconstruction performance and extended discrete codebook techniques to face image restoration. Drawing inspiration from these works, we apply the high-quality codebook prior to night image restoration.

\section{Methodology}
\subsection{Framework Overview}
To improve the recovery of high-quality images with realistic textures and normal illumination from night image $x$ containing complex degradation, we introduce a Vector-Quantized codebook as high-quality prior information to design a night image restoration network (VQCNIR). The overview of the VQCNIR framework is illustrated in Figure \ref{VQCNIR}. VQCNIR comprises an encoder $E$, an adaptive illumination enhancement module, a high-quality codebook $\mathcal{Z}$, and two decoders $G$ and $D$. Decoder G is a pre-trained decoder from VQGAN with fixed parameters. Decoder D represents the primary decoder, which progressively recovers fine details by fusing high-quality features in decoder G. 

\subsection{VQ Codebook for Priors} \label{analysis}
\textbf{VQ Codebook:} We first briefly describe the VQGAN \shortcite{VQGAN} model and its codebook, and more details can be referenced in \shortcite{VQGAN}. Given a high-quality image $x_h\in \mathbb{R}^{H\times W\times 3}$ with normal light, the encoder $E$ maps the image $x_h$ to its spatial latent representation $\hat{z}=E(x)\in \mathbb{R}^{h \times w \times n_z}$, where $n_z$ is the dimension of latent vectors. Then, each element $\hat{z}_i \in \mathbb{R}^{n_z}$ Euclidean distance nearest vector $z_k$ in the codebook is found as a VQ representation $z_\textbf{q}$ by the element-by-element quantization process $\textbf{q}(\cdot)$. It is shown as follows:
\begin{equation}
    z_\textbf{q}=\textbf{q}(\hat{z}):=\Bigg ({\text{arg} \min_{z_k\in \mathcal{Z}}||\hat{z}_i-z_k||^2_2} \Bigg)\in \mathbb{R}^{h\times w\times n_z},
\label{VQeq}
\end{equation}
where the codebook is $\mathcal{Z} = \{z_k\}_{k=1}^K \in \mathbb{R}^{K\times n_z}$ with $K$ discrete codes. Then, the decoder $G$ maps the quantized representation $z_\textbf{q}$ back into sRGB space. The overall reconstruction process can be formulated as follows:
\begin{equation}
    \hat{x}_h = G(z_\textbf{q})=G(\textbf{q}(E(x))) \approx x_h,
\end{equation}

\begin{figure}
	\begin{center}
	    \includegraphics[width=7cm]{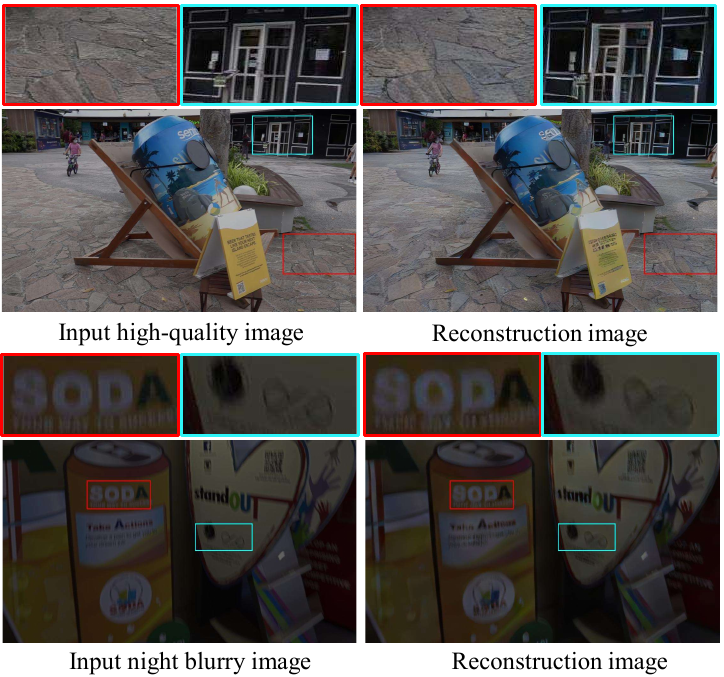}
	\end{center}
	\caption{VQGAN reconstruction results. On the left is the input image and on the right is the reconstructed image. VQGAN can provide rich detail for high-quality images but can cause some structural distortion. In degraded images, image distortion is worsened because the degraded features do not match the correct high-quality codebook features.}
	\label{VQGANAnalysis}
\end{figure}

\noindent \textbf{VQ codebook for Night Image Restoration:} To fully explore the effect of the VQ codebook prior on night image restoration, several preliminary experiments were implemented to analyze the advantages and disadvantages of VQGAN. First, we use the well-trained VQGAN to reconstruct the real image. The experimental results are shown in Figure \ref{VQGANAnalysis} top. From the figure, we can see that VQGAN can generate vivid texture details in the reconstructed images. However, some of the structural information is lost in the vector quantization process, resulting in distortion and artifacts in the reconstructed image. Therefore, the reconstruction solely depends on the quantized features in the codebook and does not yield satisfactory recovery results. The most intuitive idea is to combine the texture information generated by the quantized features from the codebook with the structural information of the latent representation to avoid structural distortion of the image.

Subsequently, we explore the effectiveness of VQGAN trained on high-quality images for the reconstruction of degraded images at night. As shown in Figure \ref{VQGANAnalysis} bottom, the restoration image is unable to recover to normal illumination due to the inconsistent illumination of the input image and the training set of VQGAN. Moreover, we found that VQGAN further deteriorates blurred textures and produces artifacts. This was attributed to the difficulty of the network in matching the correct VQ codebook features, which resulted in the inability of VQGAN under high-quality image training to recover from low illumination and blur. Therefore, we design an adaptive illumination enhancement module and a deformable bi-directional cross-attention for the mentioned low light and blur problems respectively.

\subsection{Adaptive Illumination Enhancement}

Based on the previous observations and analysis, we design an Adaptive Illumination Enhancement Module (AIEM) to solve the problem of illumination inconsistency between the quantized features and the latent features obtained from the encoder, as shown in Figure \ref{VQCNIR}. This module consists of two parts: Hierarchical Information Extraction (HIE) and Illumination Mutual Attention Enhancement (IMAE).

\noindent \textbf{Hierarchical Information Extraction:} Local lighting, such as light sources, is often observed in night-time environments. However global operation often over- or under-enhances these local regions. Thus, we employ channel attention and large kernel convolution attention to extract spatial information at different hierarchies. Specifically, HIE first employs layer normalization to stabilize the training and then performs spatial information fusion of different receptive fields. A residual shortcut is used to facilitate training convergence. Following the normalization layer, the point-wise convolution and the $3\times 3$ depth-wise convolution are used to capture spatially invariant features. Then, three parallel operators are used to aggregate channel and spatial information. The first operator uses SimpleGate \shortcite{NAFNet} to apply non-linear activation on spatially invariant features. The second operator is channel attention \shortcite{RCAN} to modulate the feature channels. The third one is the large kernel convolution attention \shortcite{LKA} to handle spatial features. The three branches output feature maps of the same size. Point-wise multiplication is used to fuse the diverse feature from the three branches directly. Finally, the output features are adjusted by point-wise convolution.

\noindent \textbf{Illumination Mutual Attention Enhancement:} According to the hierarchical information of different receptive fields obtained from the HIE, IMAE first utilize layer normalization to stabilize the training, and then illumination enhancement was applied to the features. Specifically, we design the novel illumination mutual attention convolution (IMAConv) that uses the dependencies between feature channels to estimate the curve parameters and thus adjust the illumination of the features. Two point-wise convolutions are used to adjust the input and output features of IMAConv. Residual connections are used to facilitate training convergence.

\begin{figure}
	\begin{center}
	    \includegraphics[width=7.5cm]{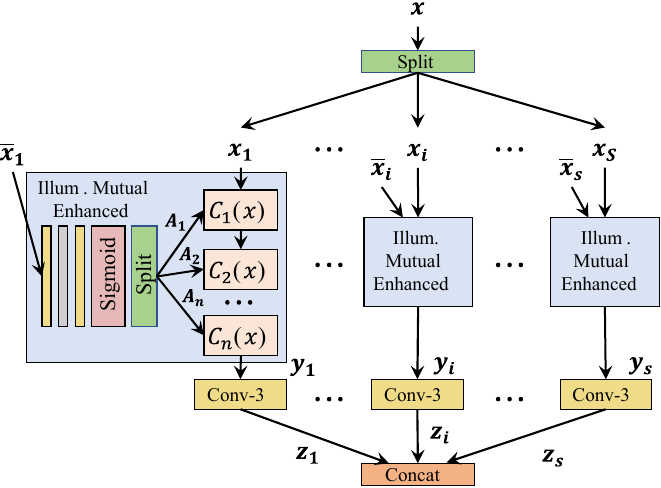}
	\end{center}
	\caption{The architecture of Illumination Mutual Attention Convolution (IMAConv).}
	\label{IMAConv}
\end{figure}

\noindent \textbf{Illumination Mutual Attention Convolution:} Considering that the illumination variation is similar between feature channels, we inspired by Zero-DCE \shortcite{Zero-DCE} introduce curve estimation and channel mutual mapping to propose an illumination mutual attention convolution that adjusts the pixel range of the feature to enhance the illumination, as shown in Figure \ref{IMAConv}. Specifically, given the input features of IMAConv as $x_f\in \mathbb{R}^{C_{in}\times H_f \times W_f}$. we first divide x into S parts at a time along the channel as follows:
\begin{equation}
    x_f^1, x_f^2, ..., x_f^S = \text{split}(x_f),
\end{equation}
where $\text{split}(\cdot)$ denotes the split operation. For each part $x_f^i\in$ $\mathbb{R}^{\frac{C_{in}}{S}\times H_f\times W_f}$, we concatenate the channel features excluding $x_f^i$ together as the complimentary to $x_i$, denoted as $\bar{x}_f^i$. Both $x_f^i$ and $\bar{x}_f^i$ are passed into the illumination mutual enhanced, which estimates multiple curve parameters $A=\{A_i\}_{i=1}^N$ through the curve estimation network $\mathcal{F}$. $A_i$ is used to adjust the range of pixel values from the features. The whole process is formulated as:
\begin{align}
    A_1, A_2, ..., A_n = \text{split}(\mathcal{F}(\bar{x_i})),\\
    y_f^i = C_n(x_f^i, A_1, A_2, \cdots A_n),
\end{align}
where $\mathcal{F}(\cdot)$ and $C_n(\cdot)$ denote the curve estimation network and the high-order curve mapping function. The curve estimation network $\mathcal{F}$ consists of three convolutional layers with kernel sizes of 5, 3, and 1, respectively, two activation functions, and a sigmoid function. For the high-order curve mapping function $C_n$, we follow the setting of Zero-DCE and adopt the following formula:
\begin{equation}
\resizebox{\columnwidth}{!}{ $
    C_n(x_f^i) = \begin{cases}
    A_1 x_f^i(1-x_f^i)+ x_f^i,  n=1  \\
        A_{n-1}C_{n-1}(x_f^i)(1-C_{n-1}(x_f^i))+C_{n-1}(x_f^i), n>1
    \end{cases}
$
}
\end{equation}
After illumination enhancement, for all $y_f^1, y_f^2, ..., y_f^S$, we use a $3\times 3$ convolution layer to generate feature $z_f^i=Conv_i(y_f^i)$. Finally, the different features $z_f^1, z_f^2, ..., z_f^S$ are concatenated to form the output of IMAConv.
\begin{equation}
    z_f = \text{Concat}(z_f^1, z_f^2, ..., z_f^S),
\end{equation}

\begin{figure}
	\begin{center}
	   \includegraphics[width=7.5cm]{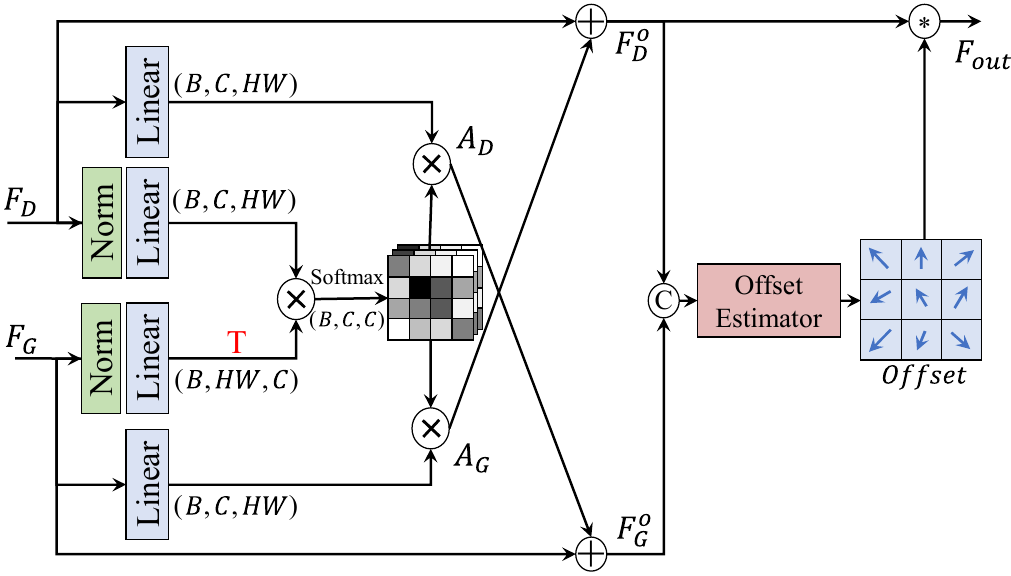}
	\end{center}
    \caption{The architecture of deformable bi-directional cross-attention (DBCA). The offset estimator module consists of a number of large kernel convolutions that use information from the large receptive fields to help fuse the features of the two decoders.}
	\label{TWAM}
\end{figure}

\subsection{Deformable Bi-directional Cross-Attention}
As previously described and analyzed, the high-quality quantized features obtained from the codebook are not flawless. The structural warping and textural distortion leads to a more severe misalignment between high-quality VQ codebook features and original degraded features. Therefore, we propose the Deformable Bi-directional Cross-Attention (DBCA) to fuse high-quality VQ codebook features and degraded features. 

Unlike the conventional cross-attention method \shortcite{crossatt}, our DBCA aims to integrate two different features using a bi-directional cross-attention mechanism and employs deformable convolutions to effectively correct the blurring degradation in the degraded feature. As shown in Figure \ref{TWAM}, given the input decoder $D$ and $G$ features $F_D$ and $F_G$, they are first mapped to corresponding $\textbf{Q}_D= W_D^p \text{LN}(F_D)$ and $\textbf{Q}_G = W_G^p \text{LN}(F_G)$ via normalization and linear layers. We further utilize linear layers to map these features to corresponding values $\textbf{V}_D$ and $\textbf{V}_G$. We reshape the aforementioned $\textbf{Q}_D$, $\textbf{Q}_G$, $\textbf{V}_D$, and $\textbf{V}_G$ into the shape of $(B, C, H*W)$ and fuse the two features using the following bi-directional cross-attention formula:
\begin{align}
    \text{A}_D &= \text{Softmax}({\textbf{Q}_D \textbf{Q}_G^T}/{\sqrt{C}}) \textbf{V}_D,\\
    \text{A}_G &= \text{Softmax}({\textbf{Q}_D \textbf{Q}_G^T}/{\sqrt{C}}) \textbf{V}_G,\\
    F_D^{o} &= \gamma_D \text{A}_G + F_D,\\
    F_G^{o} &= \gamma_G \text{A}_D + F_G,
\end{align}
where $\text{Softmax}(\cdot)$ denotes the softmax function. $\text{A}_D$ and $\text{A}_G$ respectively represent the attention maps for feature D and feature G. $\gamma_D$ and $\gamma_G$ are trainable channel-wise scales and initialized with zeros for stabilizing training.

To better fuse the high-quality codebook prior feature into the degraded feature, we first generate an offset by concatenating the two output features. Then, we use the generated offset in the deformable convolution to distort the texture feature and effectively remove the blurry degradation, which can be formalized as follows:
\begin{align}
    \text{offset} &= \text{LKConv}(\text{Concat}(F_D^o, F_G^o)),\\
    F_{out} &= \text{DeformConv}(F_D^o, \text{offset}),
\end{align}
where $\text{LKConv}(\cdot)$ and $\text{DeformConv}(\cdot)$ denote the $7\times7$ convolution and the deformable convolution, respectively.  

\subsection{Training Objectives of VQCNIR}

The training objective of VQCNIR comprises four components: (1) pixel reconstruction loss $\mathcal{L}_{pix}$ that minimizes the distance between the outputs and the ground truth; (2) code alignment loss $\mathcal{L}_{ca}$ enforces the codes of the night images to be aligned with the corresponding ground truth; (3) perceptual loss $\mathcal{L}_{per}$ which operates in the feature space, aims to enhance the perceptual quality of the restored images; and (4) adversarial loss $\mathcal{L}_{adv}$ for restoring realistic textures.

Specifically, we adopt the commonly-used L1 loss in the pixel domain as the reconstruction loss, represented by:
\begin{equation}
    \mathcal{L}_{pix} = ||x_h - VQCNIR(x_n)||_1,
\end{equation}
where the $x_h$ and $x_n$ denote high-quality ground truth and night image, respectively. To improve the matching performance of codes for night images with codes for high-quality images. We adopt the $L_2$ loss to measure the distance, which can be formulated as:
\begin{equation}
    \mathcal{L}_{ca} = ||z^e-z_\textbf{q}^e||_2^2,
\end{equation}
where $z^e$ and $z_\textbf{q}^e$ are the night image code and the ground truth code, respectively. The total training objective is the combination of the above losses:
\begin{equation}
    \mathcal{L}_{VQCNIR} = \lambda_{pix} \mathcal{L}_{pix} + \lambda_{ca} \mathcal{L}_{ca} + \lambda_{per} \mathcal{L}_{per} + \lambda_{adv} \mathcal{L}_{adv},
\end{equation}
where $\lambda_{pix}$, $\lambda_{ca}$, $\lambda_{per}$, and $\lambda_{adv}$ denote the scale factors of each loss function, respectively.

\section{Experiments}
\subsection{Dataset and Training Details}

We train our VQCNIR network on the LOL-Blur dataset \shortcite{LEDNet}, which consists of 170 sequences (10,200 pairs) of training data and 30 sequences (1,800 pairs) of test data. We use random rotations of 90, 180, 270, random flips, and random cropping to $256\times 256$ size for the augmented training data. We train our network using Adam \shortcite{Adam} optimizer with $\beta_1$=0.9, $\beta_2$=0.99 for a total of 500k iterations. The mini-batch size is set to 8. The initial learning rate is set to $1\times 10^{-4}$ and adopts the MultiStepLR to adjust the learning rate progressively. We empirically set $\lambda_{pix}$, $\lambda_{ca}$, $\lambda_{per}$, and $\lambda_{adv}$ to $\{1, 1, 1, 0.1\}$. All experiments are performed on a PC equipped with Intel Core i7-13700K CPU, 32G RAM, and the Nvidia RTX 3090 GPU with CUDA 11.2.

\begin{table}
\centering
\resizebox{8cm}{!}{%
\begin{tabular}{l|c|c|c}
\bottomrule
Method                              & PSNR$\uparrow$  & SSIM$\uparrow$  & LPIPS$\downarrow$ \\ \hline
Zero-DCE \shortcite{Zero-DCE} $\rightarrow$ MIMO \shortcite{LOLBlur_ref_7}        & 17.68 & 0.542 & 0.510 \\ 
LLFlow \shortcite{LLFlow} $\rightarrow$ Restormer \shortcite{restormer}      & 21.50 & 0.746 & 0.357 \\ 
LLFlow \shortcite{LLFlow} $\rightarrow$ Uformer \shortcite{uformer}       & 21.51 & 0.750 & 0.350 \\ \hline
MIMO \shortcite{LOLBlur_ref_7} $\rightarrow$ Zero-DCE \shortcite{Zero-DCE}        & 17.52 & 0.570 & 0.498 \\ 
Restormer \shortcite{restormer} $\rightarrow$ LLFlow \shortcite{LLFlow}     & 21.89 & 0.772 & 0.347 \\ 
Uformer \shortcite{uformer} $\rightarrow$ LLFlow  \shortcite{LLFlow}      & 21.63 & 0.758 & 0.342 \\ \hline
KinD++$^*$ \shortcite{KinD++}                             & 21.26 & 0.753 & 0.359 \\ 
DRBN$^*$  \shortcite{DRBN}                             & 21.78 & 0.768 & 0.325 \\ 
DeblurGAN-v2$^*$ \shortcite{DeblurGANv2}                       & 22.30 & 0.745 & 0.356 \\ 
DMPHN$^*$ \shortcite{LOLBlur_ref_48}                              & 22.20  & 0.817 & 0.301 \\ 
MIMO$^*$ \shortcite{LOLBlur_ref_7}                               & 22.41 & 0.835 & 0.262 \\ 
Restormer$^*$ \shortcite{restormer}                          & 23.63 & 0.841 & 0.247 \\ 
LLFlow$^*$  \shortcite{LLFlow}                            & 24.48 & 0.846 & 0.235 \\ 
LEDNet$^*$ \shortcite{LOLBlur_ref_7}                             & \underline{25.74} & \underline{0.850} & \underline{0.224} \\ \hline
Ours                                & \textbf{27.79} & \textbf{0.875} & \textbf{0.096} \\ \toprule
\end{tabular}%
}
\caption{Quantitative evaluation on the LOL-Blur dataset. The symbol $^*$ indicates the network is retrained on the LOL-Blur dataset. The best and second-best values are indicated with \textbf{Bold} text and \underline{Underline} text respectively.}
\label{tab: PSNR}
\end{table}

\begin{figure*}[t]
    \begin{center}
        \includegraphics[width=16cm]{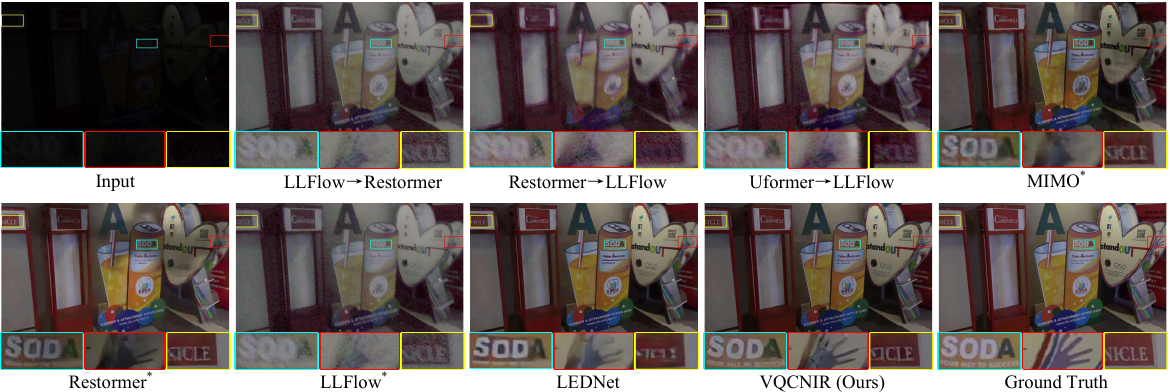}
    \end{center}
	\caption{Visual comparison results on the LOL-Blur dataset \shortcite{LEDNet}. The symbol $^*$ indicates the network is retrained on the LOL-Blur dataset. The proposed method produces visually more pleasing results. (Zoom in for the best view)}
	\label{cplolblur}
\end{figure*}

\begin{table}
\centering
\resizebox{8cm}{!}{%
\begin{tabular}{l|c|c|c}
\bottomrule
Method                      & \multicolumn{1}{l}{MUSIQ$\uparrow$} & \multicolumn{1}{l}{NRQM $\uparrow$} & \multicolumn{1}{l}{NIQE $\downarrow$} \\ \hline
RUAS \shortcite{RUAS} $\rightarrow $MIMO \shortcite{LOLBlur_ref_7}    & 34.39                               & 3.322                               & 6.812                                 \\
LLFlow \shortcite{LLFlow} $\rightarrow$ Restormer \shortcite{restormer}      & 34.45 & 5.341 & 4.803 \\ 
LLFlow \shortcite{LLFlow} $\rightarrow$ Uformer \shortcite{uformer}       & 34.32 & 5.403 & 4.941 \\ \hline
MIMO \shortcite{LOLBlur_ref_7} $\rightarrow$ Zero-DCE \shortcite{Zero-DCE} & 28.36                               & 3.697                               & 6.892                                 \\ 
Restormer \shortcite{restormer} $\rightarrow$ LLFlow \shortcite{LLFlow}     & 35.42 & 5.011 & 4.982 \\ 
Uformer \shortcite{uformer} $\rightarrow$ LLFlow  \shortcite{LLFlow}      & 34.89 & 4.933 & 5.238 \\ \hline
KinD++$^*$ \shortcite{KinD++}                 & 31.74                               & 3.854                               & 7.299                                 \\
DRBN$^*$   \shortcite{DRBN}                 & 31.27                               & 4.019                               & 7.129                                 \\
DMPHN$^*$  \shortcite{LOLBlur_ref_48}                 & 35.08                               & 4.470                               & 5.910                                 \\
MIMO$^*$ \shortcite{LOLBlur_ref_7}                   & 35.37                               & 5.140                               & 5.910                                 \\
Restormer$^*$ \shortcite{restormer}                          & 36.65 & 5.497 & 5.093 \\ 
LLFlow$^*$  \shortcite{LLFlow}                            & 34.87 & 5.312 & 5.202 \\ 
LEDNet \shortcite{LEDNet}                     & \underline{39.11}                               & \underline{5.643}                               & \underline{4.764}                                 \\ \hline
Ours                        & \textbf{51.04}                               & \textbf{7.064}                               & \textbf{4.599}                                 \\ \toprule
\end{tabular}%
}
\caption{Quantitative evaluation on the Real-LOL-Blur dataset. The symbol $^*$ indicates the network is retrained on the LOL-Blur dataset. The best and second-best values are indicated with \textbf{Bold} text and \underline{Underline} text respectively.}
\label{tab: MUSIQ}
\end{table}

\subsection{Results on LOL-Blur dataset}
In this section, we compare our proposed VQCNIR quantitatively and qualitatively with all the above methods on the LOL-Blur test set \shortcite{LEDNet}. We use the two most widely evaluated metrics: PSNR and SSIM for a fair evaluation of all methods. In addition, we employ the LPIPS metric to evaluate the perceptual quality of the restored images.

\noindent\textbf{Quantitative Evaluations.} Table \ref{tab: PSNR} shows the quantitative results of our method and other methods on the LOL-Blur dataset. As can be seen from the table, our method can outperform the most superior LEDNet method by 2.05 dB and 0.025 in terms of PSNR and SSIM metrics. Besides, our method is far superior to existing methods in terms of perceptual metrics. These results provide sufficient evidence of the effectiveness of our method.

\noindent\textbf{Qualitative Evaluations.} Figure \ref{cplolblur} shows the visual effect of all the compared methods. As the figure shows, most methods are ineffective in removing the blurring effect in severely blurred regions, inevitably introducing artifacts into the restored image. In contrast, our method can effectively recover the correct texture features by using high-quality prior information. Therefore, these results provide sufficient evidence that the codebook prior proposed by our method is particularly suitable for the task of night image restoration. More results are provided in the supplementary material.

\begin{figure*}[t]
	\begin{center}
	    \includegraphics[width=16cm]{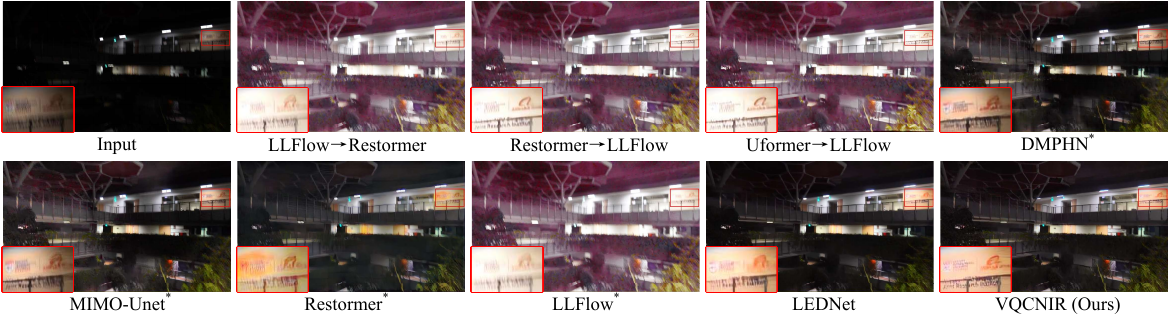}
	\end{center}
	\caption{Visual comparison on the Real-LOL-Blur dataset \shortcite{LEDNet}. The symbol $^*$ indicates the network is retrained on the LOL-Blur dataset. The proposed method produces visually more pleasing results. (Zoom in for the best view)} 
	\label{cpreallol}
\end{figure*}

\subsection{Results on Real dataset}
To better illustrate the effectiveness of our method in the real scene, we compare our proposed VQCNIR with the above method quantitatively and qualitatively under the real Real-LOL-Blur dataset \shortcite{LEDNet}. Since the real scene lacks a corresponding reference image to evaluate, three non-reference evaluation metrics were used for the evaluation: MUSIQ \shortcite{MUSIQ}, NRQM \shortcite{NRQM}, and NIQE \shortcite{NIQE}. The MUSIQ metric assesses mainly color contrast and sharpness, which is more appropriate for this task. 

\noindent\textbf{Quantitative Evaluations.} 
Table \ref{tab: MUSIQ} exhibits the quantitative results of our method and other methods on the Real-LOL-Blur test set. As shown in Table \ref{tab: MUSIQ}, our method achieves the highest NIQE and NRQM scores, indicating that the restored results of our method have better image quality and are consistent with human perception. Moreover, we have the highest MUSIQ, which means that our results are the best in terms of color contrast and sharpness.

\noindent\textbf{Qualitative Evaluations.}
Figure \ref{cpreallol} displays the visual comparison results for all evaluated methods. As evident from the figure, simple cascade deblurring and low-light enhancement techniques can cause issues such as overexposure and blurring of saturated areas in the image. Even the end-to-end method of retraining on the LOL-Blur dataset suffers from undesired severe artifacts and blurring. In contrast, our proposed VQCNIR outperforms these methods in terms of visual quality, demonstrating fewer artifacts and blurring. This improvement can be attributed to the successful integration of a high-quality codebook prior into the network, which assists in generating high-quality textures. The comparison results of a real-world image further demonstrate the superiority of our proposed method. More results are provided in the supplementary material.

\begin{table}[t]
\begin{center}
\resizebox{8cm}{!}{%
\begin{tabular}{l|ccc|cc}
\hline
\multicolumn{1}{c|}{\multirow{2}{*}{Models}} & \multicolumn{3}{c|}{Configuration} & \multicolumn{2}{c}{LOL-Blur} \\
\multicolumn{1}{c|}{}                        & Decoder D    & AIEM    & DBCA    & PSNR    & SSIM     \\ \hline
VQGAN                                        &                &         &         & 10.79   & 0.3028   \\
Setting 1                                    & \ding{52}            &         &         & 26.58   & 0.8486    \\
Setting 2                                    & \ding{52}            &         & \ding{52}     & 26.89   & 0.8599   \\
Setting 3                                    & \ding{52}            & \ding{52}     &         & 27.48   & 0.8692  \\ \hline
VQCNIR                                       & \ding{52}            & \ding{52}     & \ding{52}     & 27.79   & 0.8750  \\ \hline
\end{tabular}%
}
\end{center}
\caption{Ablation studies of different components. We report the PSNR and SSIM values on the LOL-Blur dataset.}
\label{tab:ablationmodel}
\end{table}

\subsection{Ablation Study}
In this section, we have implemented a series of ablation experiments to better validate the effectiveness of each of our proposed modules. To verify the effectiveness of our proposed operations, a series of ablation experiments are presented and the results are shown in Table \ref{tab:ablationmodel}. Initially, we use the VQGAN as our baseline model. Table \ref{tab:ablationmodel} shows that VQGAN does not effectively address low light and blur degradation, since VQGAN is a codebook prior learned from high-quality natural images and is unable to correctly match degraded features. By designing corresponding parallel decoders, the network can then effectively use high-quality priors to assist in the reconstruction of degraded features. However, the illumination inconsistency between the degraded features and the codebook prior can prevent accurate matching of the high-quality prior features, leading to the occurrence of artifacts. Furthermore, the degraded features are at some distance from high-quality features. Therefore, AIEM and DBCA can be used to effectively improve network performance and image quality. More results of the ablation experiments are provided in the supplementary material.

\section{Conclusion}

In this work, we introduce high-quality codebook priors and propose a new paradigm for night image restoration called VQCNIR. Through analysis, we discover that directly applying codebook priors can result in improper matching between degraded features and high-quality codebook features. To address this, we propose an Adaptive Illumination Enhancement Module (AIEM) and a Deformable Bi-directional Cross-Attention (DBCA) module, leveraging estimated illumination curves and bi-directional cross-attention. By fusing codebook priors and degraded features, VQCNIR effectively restores normal illumination and texture details from night images. Extensive experiments demonstrate the state-of-the-art performance of our method.

\section*{Acknowledgments}
This work was supported by the Science and Technology Project of Guangzhou under Grant 202103010003, Science and Technology Project in key areas of Foshan under Grant 2020001006285.

\bibliography{aaai24}

\end{document}